\def\year{2021}\relax
\documentclass[letterpaper]{article} 
\usepackage{aaai21}  
\usepackage{times}  
\usepackage{helvet} 
\usepackage{courier}  
\usepackage[hyphens]{url}  
\usepackage{graphicx} 
\usepackage{graphicx}  
\usepackage{natbib}  
\usepackage{caption2} 
\nocopyright
\title{Self-supervised Clipped-objective Reinforcement Leaning Framework for Risk-aware Portfolio Management}
\author{
    Zongshen Mu\textsuperscript{\rm 1}\\
 }
\affiliations{
    \textsuperscript{\rm 1}Southwest Securities Co., Ltd., Chongqing, China\\
    drzsmu@163.com
}

\usepackage[colorlinks, breaklinks = true]{hyperref}		
\usepackage{amsmath}
\usepackage{amssymb}
\usepackage{algorithm}
\usepackage{algpseudocode}
\usepackage{booktabs}
\usepackage{graphicx}
\usepackage{subfigure}
\usepackage{booktabs}
\usepackage{enumitem}
\usepackage{multirow}
\usepackage{stfloats}
\usepackage{setspace}

\begin{document}

\nocopyright

\maketitle

\begin{abstract}
Portfolio management involves balancing diverse financial assets, a critical yet challenging task in quantitative trading. Although deep reinforcement learning has emerged as an active research domain for solving complex sequential decision-making problems over the past decade, deploying reinforcement learning frameworks in non-stationary financial markets raises formidable difficulties . Most existing reinforcement learning-based portfolio management models fail to account for tick-level intraday market dynamics (e.g., order flow imbalances and liquidity fluctuations), which reflect rapid capital movements . Simultaneously, these models suffer from unstable learning processes due to reinforcement learning’s inherent high variance . This paper presents a \textbf{S}elf-supervised \textbf{C}lipped-objective \textbf{P}ortfolio management framework (SCP), which addresses these limitations through a two-phase architecture. Market feature integration phase (application contribution) integrates micro-level market microstructure patterns with macro-level financial indicators to enhance sensitivity to intraday asset trend dynamics. Policy optimization Phase (AI contribution) employs intermediate asset state representations to derive optimal portfolio allocations via a self-supervised clipped-objective policy-learner, ensuring both effectiveness and computational efficiency. Extensive evaluations on three major index datasets from China's A-share market demonstrate that SCP's superior performance in balancing risk-return trade-offs compared to state-of-the-art baselines. Exploratory and ablation studies further validate the contributions of SCP's key components.

\end{abstract}

\section{Introduction}
Portfolio management constitutes a fundamental quantitative trading task where investors dynamically allocate capital across diverse financial assets, periodically adjusting portfolio weights to maximize risk-adjusted returns within predefined risk tolerance thresholds. This discipline, alternatively termed portfolio optimization or strategic asset allocation in academic literature, faces persistent challenges in balancing predictive accuracy with computational tractability. 

The Markowitz Modern Portfolio Theory (MPT) framework~\cite{markowitz2009harry,steinbach2001markowitz} pioneers mathematical portfolio construction through mean-variance analysis, yet suffers from two inherent limitations: 1) dependency on precise asset return predictions that prove elusive given financial markets' high noise-to-signal characteristics , and 2) restrictive assumptions about return distributions that poorly align with real-world asset price dynamics . Recent advances~\cite{hambly2023recent,li2000optimal,mossin1968optimal} highlight how limited data availability and non-stationary market behaviors exacerbate these theoretical constraints.  

The Markowitz Modern Portfolio Theory (MPT) framework~\cite{markowitz2009harry,steinbach2001markowitz} establishes the mathematical foundation for optimal asset allocation through mean-variance optimization, yet its practical application remains constrained by fundamental theoretical premises. The model relies on precise asset return predictions that prove elusive given financial markets' high noise-to-signal characteristics. Moreover, restrictive assumptions about return distributions poorly align with real-world asset price dynamics. Contemporary studies~\cite{hambly2023recent,li2000optimal,mossin1968optimal} further demonstrate how non-stationary market regimes and path-dependent price dynamics exacerbate these limitations, as historical covariance estimates derived from finite datasets provide unreliable guidance for forward-looking portfolio construction.

Recently, deep learning architectures~\cite{zou2022stock} demonstrate proficiency in price forecasting through supervised approaches~\cite{ke2017lightgbm,feng2019temporal,ding2020hierarchical,ding2015deep,hu2018listening,xu2018stock,zhang2017stock}. Though such methods exhibit critical shortcomings by decoupling prediction from portfolio optimization. Their singular focus on price trajectory estimation neglects essential risk dimensions and cross-asset correlation dynamics, often yielding suboptimal allocation strategies. 

Deep reinforcement learning (DRL) emerges as a promising alternative by formalizing portfolio management as sequential decision-making under uncertainty~\cite{sun2023reinforcement}. Despite notable progress in DRL applications~\cite{deng2016deep,ding2018investor,jin2016portfolio,wang2019alphastock,ye2020reinforcement,wang2021deeptrader,niu2022metatrader}, two critical limitations persist: First, existing approaches predominantly operate at daily temporal resolutions, failing to exploit intraday microstructure patterns that capture ephemeral arbitrage opportunities~\cite{sun2022deepscalper}. Second, reliance on on-policy optimization mechanisms~\cite{hausknecht2016policy} introduces computational inefficiencies due to the inherent exploration-exploitation tradeoff, particularly problematic in high-frequency trading contexts where sample efficiency determines profitability.

This paper presents SCP, a novel self-supervised clipped-objective reinforcement learning framework for portfolio management that decouples the optimization process into two synergistic components. The first module addresses heterogeneous data integration through parallel processing of macro-level economic indicators and micro-level asset dynamics, where dual-stream Transformer architectures - comprising time-series and graph-based encoders - jointly analyze low-frequency market trends and high-frequency trading signals. This multimodal fusion generates asset embeddings encapsulating spatiotemporal patterns predictive of future price movements. The subsequent module implements stable off-policy optimization through trajectory-aware variance reduction, employing clipped-objective learning to enable conservative yet effective policy updates. A novel trajectory ranking buffer prioritizes high-reward action sequences while the clipped loss function ensures controlled gradient steps, effectively balancing exploration robustness with update efficiency.

In a nutshell, our principal contributions are threefold:
\begin{itemize}
	\item We pioneer the simultaneous integration of daily and intraday market data through adaptive frequency encoding, enhancing cross-scale pattern recognition for opportunistic asset allocation.
	\item We develop a self-supervised clipped optimization paradigm that mitigates premature convergence risks through reward-aware sample prioritization and gradient clipping mechanisms.
	\item Extensive empirical validation across three major Chinese A-share indices establishes the framework's superiority over state-of-the-art MetaTrader-based benchmarks, with statistically significant enhancements of approximately 20\% in risk-adjusted returns (ASR) and 30\% reduction in maximum capital erosion (MDD).
\end{itemize}

\section{Related Work}
\subsection{Traditional Portfolio Investment Strategies}
Three classical approaches dominate traditional portfolio management: momentum trading, mean reversion, and multi-factor models. The momentum trading strategy~\cite{jegadeesh2002cross} operates under the assumption that historical price trends tend to persist, implementing a "buy winners, sell losers" approach by purchasing assets with strong historical performance and divesting underperforming ones. The mean reversion strategy~\cite{efficiency1993returns} follows a contrarian philosophy, acquiring assets trading below their historical average price levels and liquidating positions when prices revert to or exceed their long-term mean. Multi-factor models~\cite{fama1996multifactor} employ a systematic framework that combines multiple economic indicators through statistical optimization to evaluate and select securities. However, these traditional approaches generally lack the temporal resolution to capture intraday market dynamics and are effectiveness primarily in specific market regimes. Recent empirical evidence suggests that machine learning algorithms have achieved significant performance improvements over these conventional methodologies, particularly in processing high-frequency financial data and adapting to complex market conditions.

\subsection{Deep Supervised Learning in Portfolio Management}
Capitalizing on their capability to extract hierarchical representations from large-scale datasets, deep neural networks have become pivotal in data-driven asset allocation strategies. Supervised learning frameworks typically adopt two methodological paradigms: return prediction through regression analysis, and price trend classification via binary or multi-class categorization. These predictive outputs are systematically transformed into trading decisions through rule-based mechanisms, such as top-k selection or threshold-based position adjustments. Substantial research efforts have focused on enhancing predictive accuracy through architectural innovations in temporal sequence modeling and multimodal feature fusion, as demonstrated in~\cite{hou2021stock,hu2018listening,qin2017dual,yoo2021accurate}. Nevertheless, these supervised learning approaches decouple prediction and decision-making phases, preventing direct interaction with market feedback loops and leading to suboptimal adaptation to evolving market conditions.

\begin{figure*}[!htp]
	\centering
	\includegraphics[width=1\linewidth]{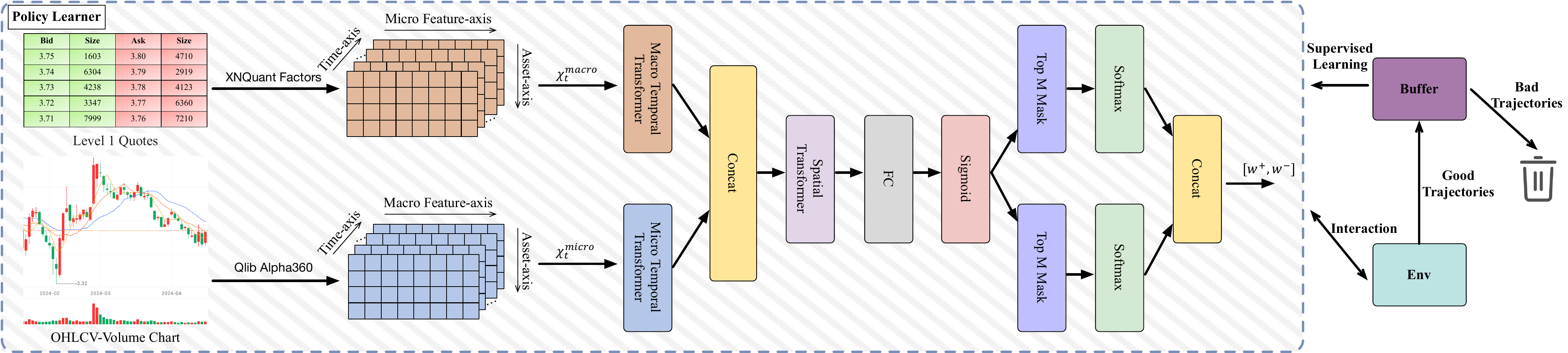}
	\vspace{-20pt}
	\caption{The proposed network structure for portfolio generation.}
	\vspace{-20pt}
	\label{fig_ramework}
	\
\end{figure*}

\subsection{Deep Reinforcement Learning in Portfolio Management}
Deep reinforcement learning (DRL) in portfolio management primarily adopts two methodological paradigms: value-based approaches that estimate state-action utilities through Q-learning variants, and policy-based methods that directly optimize allocation strategies. While value-based RL learns value functions to evaluate the expected returns of trading actions in dynamic markets~\cite{jin2016portfolio,fischer2018reinforcement,guidotti2018survey}, its effectiveness is constrained by the inherent complexity of financial environments where precise value approximation becomes intractable. This limitation drives the dominance of policy-based methods, which treat portfolio weights as direct policy outputs and leverage neural networks to parameterize allocation strategies through gradient ascent.

Recurrent Reinforcement Learning (RRL)~\cite{moody1998performance}, as a foundational policy-based algorithm, introduces a differential Sharpe ratio optimization mechanism to maximize cumulative portfolio returns while considering transaction costs. The FDDR model~\cite{deng2016deep} enhances the RRL framework by integrating deep neural networks for temporal pattern extraction. Investor-Imitator~\cite{ding2018investor} formalizes trading knowledge through logic-constrained action mapping, synthesizing investor behaviors from historical trading sequences and expert descriptors. EIIE~\cite{jiang2017deep} integrates Deep Deterministic Policy Gradient (DDPG) architecture with hierarchical feature aggregation, combining multi-scale market representations from different neural network layers through attention mechanisms. SARL~\cite{ye2020reinforcement} extends the EIIE framework by incorporating heterogeneous data sources including market sentiment indicators and volatility indices, thereby enhancing environmental adaptability. Graph-based approaches like Alphastock~\cite{wang2019alphastock} and LSRE-CAAN~\cite{li2023online} leverage industry correlation graphs and causal relationship mining to mitigate asset allocation bias through spatial dependency modeling. For risk-return balancing, DeepTrader~\cite{wang2021deeptrader} implements dynamic long-short allocation strategies conditioned on macro market indicators such as volatility indices and yield curves. MetaTrader~\cite{niu2022metatrader} achieves policy adaptability through imitation learning-guided exploration and meta-learning-based market regime switching detection. High-frequency trading innovations include DeepScalper~\cite{sun2022deepscalper}, which emulates professional traders' order execution patterns through tick-level data processing and micro-price impact modeling. MarginTrader~\cite{gu2023margin} evolves from cash-only to margin trading paradigms by incorporating real-world constraints including short-selling mechanics and regulatory capital requirements. QFPIS~\cite{qiu2024design} introduces dual-agent decision architecture, where one policy network generates portfolio weights while another determines asset holding durations through portfolio rebalancing thresholds.

Recent reinforcement learning algorithms like DPO (Direct Preference Optimization)~\cite{rafailov2024direct} and GRPO (Group Relative Policy Optimization)~\cite{shao2024deepseekmath} have gained significant traction, largely inspired by the rise of large language models (LLMs). For portfolio optimization, however, these methods require manually curated high-quality investment portfolio data, which can be costly and resource-intensive for model training. Consequently, we opt for the Proximal Policy Optimization (PPO) algorithm to heuristically learn portfolio strategies based on historical factor data and custom-designed reward functions.

Although deep reinforcement learning (DRL) algorithms have achieved remarkable capabilities, their training process remains notoriously unstable due to fundamental challenges, including policy gradient variance amplification and value function overestimation. Our framework addresses the state-action distribution shift through an iterative process: (1) collecting high-reward trajectories via prioritized exploration, and (2) performing imitation learning with contrastive ranking losses in the experience buffer. Building upon proximal policy optimization (PPO)~\cite{schulman2017proximal} which enhances sample efficiency through multi-epoch minibatch updates, we further integrate self-supervised representation learning to stabilize policy iteration. To prevent premature convergence, we derive alternative policy gradient estimators with variance-reduction mechanisms that maintain exploration capacity during optimization.

\section{Problem Setup}
\subsection{Problem Formulation}
Portfolio management is a sequential decision-making process that dynamically allocates capital across assets under market uncertainty to optimize risk-adjusted returns. This problem is formally modeled as a Markov Decision Process (MDP) $M=(\mathcal{S},\mathcal{A},\mathcal{P},\mathcal{R})$ with the following components:

\textbf{State Space $\mathcal{S}$}: The composite state $s_t = \{X^{macro}_t, X^{micro}_t\} \in \mathcal{S}$ integrates multi-frequency market signals. $X^{macro}_t \in \mathbb{R}^{d_1 \times L}$ encodes low-frequency features (price-volume trends, fundamental indicators) over $L$ historical periods. $X^{micro}_t \in \mathbb{R}^{d_2 \times H}$ captures high-frequency order book dynamics (bid-ask spreads, market depth) within $H$ intraday intervals. $d_1$, $d_2$ denote feature dimensions and the look-back window $L,H$ is user-configurable.

\textbf{Action Space $\mathcal{A}$}: The portfolio allocation vector $a_t = [\boldsymbol{w}^+_t; \boldsymbol{w}^-_t] \in \mathcal{A}$ satisfies:
\begin{equation}
    \begin{cases}
        \sum_{i=1}^N w^+_{t,i} = 1, & w^+_{t,i} \in [0,1] \ \text{(long positions)} \\
        \sum_{i=1}^N w^-_{t,i} = -1, & w^-_{t,i} \in [-1,0] \ \text{(short positions)}
    \end{cases}
\end{equation}
where $N$ denotes the asset universe size. Practical implementations typically select top-ranked positions through thresholding operations.

\textbf{Transition Dynamics $\mathcal{P}$}: The state evolution $P(s_{t+1}|s_t,a_t)$ follows market microstructure principles under the price-taker assumption, where individual trades negligibly impact market equilibrium prices.

\textbf{Reward Function $\mathcal{R}$}: The differential Sharpe ratio~\cite{moody1998performance} serves as the immediate reward:
\begin{equation}
    r_t = \frac{\mu_{p,t} - r_f}{\sigma_{p,t}}  \label{eq_sharpe}
\end{equation}
where $\mu_{p,t}$ denotes portfolio return, $r_f$ the risk-free rate (like that of government bonds), and $\sigma_{p,t}$ return volatility computed over a rolling window.

\textbf{Policy Optimization}: The agent's policy $\pi_\theta(a_t|s_t)$ parameterized by $\theta$ selects an action $a_t$ for a given state $s_t$. When the action $a_t$ is executed and interacted with the environment, the state is changed to $s_{t+1}$ by the transition function $P(s_{t+1}|s_t,a_t)$. It aims to maximize the expected cumulative risk-adjusted return (Sharpe ratio):
\begin{equation}
    \max_\theta \mathbb{E}_{\pi_\theta}\left[\sum_{t=1}^T \gamma^{t-1} r_t\right]
\end{equation}
where $\gamma \in [0,1)$ is the discount factor governing myopic versus far-sighted strategies

\subsection{Trading Procedure}

We adopt a long-short strategy under the following assumptions: (1) all assets permit both long and short positions without liquidity constraints, and (2) the investor maintains sufficient capital reserves for dynamic rebalancing operations. At each rebalancing interval $t$, a reinforcement learning (RL) agent generates normalized portfolio weights $[\boldsymbol{w}^+_t, \boldsymbol{w}^-_t]$ based on multi-frequency market state representations. Here, $\boldsymbol{w}^+_t \in \mathbb{R}^N$ denotes the long exposure vector capturing upside potential through price appreciation $\boldsymbol{r}^+_t$, while $\boldsymbol{w}^-_t \in \mathbb{R}^N$ represents the short exposure vector profiting from price depreciation $\boldsymbol{r}^-_t$. The net portfolio return at period $t$ is computed as:  
\begin{equation}
\mu_{p,t} = \boldsymbol{w}^+_t \cdot \boldsymbol{r}^+_t - \boldsymbol{w}^-_t \cdot \boldsymbol{r}^-_t
\end{equation}
Following this rule, the RL framework dynamically adjusts multi-asset exposures to maximize cumulative risk-adjusted returns. This formulation aligns with modern portfolio theory's emphasis on balancing return generation and risk mitigation through strategic diversification.

\section{Methodology}
Figure \ref{fig_ramework} shows an overview of a self-supervised clipped-objective RL-based portfolio management framework. It contains two components, first is a policy learner based on macro and micro features of assets, and then is self-supervised clipped-objective policy optimization. There are dual temporal Transformer branches and joint spatial Transformer for asset representation learning and MLPs for portfolio generation in the policy learner.

\subsection{Feature Extraction}
The input state representation $s_t$ integrates multi-frequency market signals through a hierarchical processing pipeline: macro-level daily indicators ${X}^{macro}_t$ and micro-level intraday indicators ${X}^{micro}_t$ first undergo parallel processing through dedicated temporal Transformers. These domain-specific encoders extract multi-scale temporal dependencies - the macro Transformer focuses on long-term economic cycles and policy impacts over historical days, while the micro Transformer models short-term price momentum and liquidity patterns across intraday intervals. The encoded temporal features then feed into a cross-asset spatial Transformer that learns dynamic interaction weights between financial instruments through self-attention mechanisms, quantifying how sector-specific news sentiment propagates across industries and how market shocks create correlated asset movements. The resultant integrated representation, which jointly encapsulates temporal evolution patterns and spatial correlation structures across assets, is subsequently passed to the portfolio generator for allocation decision-making.

Let ${X}^{macro}_{t,i} = (\boldsymbol{x}^{ma}_{t-T',i}, ..., \boldsymbol{x}^{ma}_{t,i}) \in \mathbb{R}^{T' \times d}$ denote the historical daily features for asset $i$ over $T'$ look-back periods, where $\boldsymbol{x}^{ma}_{t,i} \in \mathbb{R}^d$ represents Qlib Alpha360-processed daily OHLCV (open/high/low/close price and volume) metrics~\cite{yang2020qlib} mapped through nonlinear transformations. Positional embeddings $\boldsymbol{p}^{ma}_{t,i} \in \mathbb{R}^d$ are generated via sinusoidal encoding:
\begin{equation}
p^{ma}_{t,i,2m} = \sin(j/10000^{2m/d}), \quad  p^{ma}_{t,i,2m+1} = \cos(j/10000^{2m/d})
\end{equation}
where $m$ indexes embedding dimensions and $j$ denotes temporal position. The macro-level Transformer encoder combines input features ${X}^{macro}_t$ with positional embeddings ${P}^{macro}_t$ through element-wise addition ${X'}^{macro}_t = {X}^{macro}_t \oplus {P}^{macro}_t$, then processes them via stacked identical layers. a multi-head attention sub-layer~\cite{vaswani2017attention} and an MLP sub-layer. Each sub-layer is followed by residual connection and layer normalization:
\begin{align}
H^{Macro}_t &= \text{LayerNorm}\left({X'}^{Macro}_t + \text{MultiHead}({X'}^{Macro}_t)\right) \\
L^{Macro}_t &= \text{LayerNorm}\left(H^{Macro}_t + \text{MLP}(H^{Macro}_t)\right)
\end{align}
where LayerNorm is layer normalization~\cite{ba2016layer}, MLP refers to a two-layer feed-forward network with ReLU activation function, MultiHead denotes the multi-head attention mechanism.

The multi-head mechanism computes $h$ parallel attention heads:
\begin{align}
	&\text{MultiHead}({X'}^{Macro}_t)=||^{h}_{i=1}(\text{head}_i)W^O \label{eq_multi}\\
	&\text{head}_{i} = \text{Attention}(Q_i,K_i,V_i)=\text{softmax}(\frac{Q_iK^{\top}_i}{\sqrt{d/h}}V_i) \label{eq_head}
\end{align}
where $||$ is the concatenation operation, $Q_i = {X'}^{Macro}_tW^Q_i$, $K_i = {X'}^{Macro}_tW^K_i$, $V_i = {X'}^{Macro}_tW^V_i$ with learnable parameters $W^Q_i, W^K_i, W^V_i \in \mathbb{R}^{d \times d/h}$ and $W^O \in \mathbb{R}^{d \times d}$. This architecture enables parallel computation of long-range dependencies while avoiding gradient vanishing/explosion issues prevalent in RNN-based models.

Let ${X}^{micro}_{t,i} = (\boldsymbol{x}^{mi}_{t-T'',i}, ..., \boldsymbol{x}^{mi}_{t,i}) \in \mathbb{R}^{T'' \times d}$ represent the high-frequency intraday time-series data for asset $i$ over $T''$ look-back intervals, where $\boldsymbol{x}^{mi}_{t,i} \in \mathbb{R}^d$ encapsulates order book dynamics including bid-ask spreads, market depth, and derived statistical measures (kurtosis, skewness, extreme value differentials). The micro-level Transformer encoder mirrors the architectural design of its macro-level counterpart:
\begin{align}
H^{Micro}_t &= \text{LayerNorm}\left({X'}^{Micro}_t + \text{MultiHead}({X'}^{Micro}_t)\right) \\
L^{Micro}_t &= \text{LayerNorm}\left(H^{Micro}_t + \text{MLP}(H^{Micro}_t)\right)
\end{align}
where ${X'}^{Micro}_t = {X}^{Micro}_t \oplus {P}^{Micro}_t$ combines raw micro-features with positional embeddings via element-wise addition. 

The final asset representation $\boldsymbol{e}_{t,i} = L^{Macro}_t|| L^{Micro}_t$ integrates multi-scale temporal patterns, enabling simultaneous modeling of macroeconomic trends and microstructure effects. This dual-scale encoding mechanism advances beyond conventional single-scale representations by explicitly capturing high-frequency liquidity patterns and transient market anomalies - critical drivers of short-term price dynamics often overlooked in prior work.

%

The industry lead-lag effect hypothesis posits that cross-asset return correlations exhibit directional propagation patterns, where certain stocks systematically lead or lag peers within sectoral clusters~\cite{ye2021multi}. To operationalize this phenomenon, we construct a sector-attributed graph ${G}=({V},{E})$ where nodes ${V}=\{\boldsymbol{e}_{t,i}\}^n_{i=1}$ represent assets encoded with multi-scale state embeddings, and edges ${E}$ capture inter-asset relationships guided by industry classifications. For asset $i$, its neighborhood $\mathcal{N}_i = \{\boldsymbol{e}_{t,j} : (i,j) \in {E}\}$ defines sectorally-aligned peers exhibiting statistically significant lead-lag correlations. Modern graph neural networks~\cite{hamilton2017inductive} implement this inductive bias through iterative neighborhood aggregation mechanisms. At layer $l$, each node's representation $\boldsymbol{h}^{(l)}_{t,i}$ updates via:  
\begin{align}
\boldsymbol{a}^{(l)}_{t,i} &= \text{AGGREGATE}^{(l)}\left(\{\boldsymbol{h}^{(l-1)}_{t,j} : j \in \mathcal{N}_i\}\right) \label{agg_eq} \\
\boldsymbol{h}^{(l)}_{t,i} &= \text{COMBINE}\left(\boldsymbol{h}^{(l-1)}_{t,i}, \boldsymbol{a}^{(l)}_{t,i}\right)
\end{align}  
with $\boldsymbol{h}^{(0)}_{t,i} = \boldsymbol{e}_{t,i}$. Through $L$ hierarchical aggregation steps, the final node embedding $\boldsymbol{h}^{(L)}_{t,i}$ encapsulates both asset-specific dynamics and $L$-hop sectoral influence patterns. This contrasts with conventional dense correlation graphs that introduce noise through arbitrary pairwise connections, instead leveraging sector taxonomy to focus on economically-grounded interactions.

Recent advances in graph neural architectures introduce diverse implementations for $\text{AGGREGATE}^{(l)}(\cdot)$ and $\text{COMBINE}^{(l)}(\cdot)$ functions. Our implementation adopts the masked transformer convolution ($\text{TransformerConv}$) from UniMP~\cite{shi2020masked}, where the aggregation operation is defined as:
\begin{align}
\boldsymbol{a}^{(l)}_{t,i} &= ||^{C}_{c=1}[\sum\nolimits_{j\in \mathcal{N}(i)}\alpha^{(l)}_{c,t,ij}*\boldsymbol{v}^{(l)}_{c,t,j}] \\
\alpha^{(l)}_{c,t,ij} &= \frac{\exp\left(\langle\boldsymbol{q}^{(l)}_{c,t,i},\boldsymbol{k}^{(l)}_{c,t,j}\rangle/\sqrt{d_c}\right)}{\sum_{k\in\mathcal{N}_i}\exp\left(\langle\boldsymbol{q}^{(l)}_{c,t,i},\boldsymbol{k}^{(l)}_{c,t,k}\rangle/\sqrt{d_c}\right)}
\end{align}
where $\langle\boldsymbol{q},\boldsymbol{k}\rangle = \boldsymbol{q}^\top\boldsymbol{k}$ denotes the dot-product similarity, and $d_c$ is the dimension per attention head. The query $\boldsymbol{q}^{(l)}_{c,t,i}$, key $\boldsymbol{k}^{(l)}_{c,t,j}$, and value $\boldsymbol{v}^{(l)}_{c,t,j}$ vectors are derived through linear transformations analogous to Equations~\ref{eq_multi}-\ref{eq_head} in the temporal Transformer. The combination function incorporates gated residual connections to mitigate over-smoothing:
\begin{align}
\boldsymbol{h}^{(l)}_{t,i} &= \text{ReLU}\left(\text{LayerNorm}\left((1-\beta^{(l)}_{t,i})\boldsymbol{a}^{(l)}_{t,i} + \beta^{(l)}_{t,i}\boldsymbol{r}^{(l)}_{t,i}\right)\right) \\
\boldsymbol{r}^{(l)}_{t,i} &= W^{(l)}_r\boldsymbol{h}^{(l-1)}_{t,i} + \boldsymbol{b}^{(l)}_r \\
\beta^{(l)}_{t,i} &= \sigma\left(W^{(l)}_g[\boldsymbol{a}^{(l)}_{t,i};\boldsymbol{r}^{(l)}_{t,i};\boldsymbol{a}^{(l)}_{t,i}-\boldsymbol{r}^{(l)}_{t,i}]\right)
\end{align}
where $\sigma(\cdot)$ denotes the sigmoid function. In the final GNN layer ($l=L$), we employ head averaging and remove nonlinear activations:
\begin{align}
\boldsymbol{a}^{(L)}_{t,i} &= \frac{1}{C}\sum_{c=1}^C\sum_{j\in\mathcal{N}_i}\alpha^{(L)}_{c,t,ij}\boldsymbol{v}^{(L)}_{c,t,j} \\
\boldsymbol{h}^{(L)}_{t,i} &= (1-\beta^{(L)}_{t,i})\boldsymbol{a}^{(L)}_{t,i} + \beta^{(L)}_{t,i}\boldsymbol{r}^{(L)}_{t,i}
\end{align}

\subsection{Portfolio Generator}
The encoded representation $\boldsymbol{h}^{(L)}_{t,i}$ for asset $i$ undergoes dimensionality reduction and score transformation through a feed-forward layer with sigmoid activation:  
\begin{equation}
v_{t,i} = 2\cdot\sigma(W_p\boldsymbol{h}^{(L)}_{t,i} + b_p) - 1  
\end{equation}  
where $W_p$, $b_p$ are learnable parameters, scaling the ranking score $v_{t,i} \in [-1,1]$.  

The portfolio allocation mechanism proceeds as follows. Rank all assets in descending order based on $\boldsymbol{v}_t = [v_{t,1},...,v_{t,N}]$. Then construct long portfolios $\boldsymbol{V}_t^+$ (Top-$M$ assets $w^+_{t,i}>0$) and short portfolios $\boldsymbol{V}_t^-$ (Bottom-$G$ assets $w^-_{t,i}<0$). The normalized weights are computed via partitioned softmax:  
\begin{equation}
w^+_{t,i} = \begin{cases} 
\frac{\exp(v_{t,i})}{\sum_{j\in\boldsymbol{V}_t^+}\exp(v_{t,j})} & i \in \boldsymbol{V}_t^+ \\ 
0 & \text{otherwise}
\end{cases}, \quad  
w^-_{t,i} = \begin{cases} 
\frac{\exp(1-v_{t,i})}{\sum_{j\in\boldsymbol{V}_t^-}\exp(1-v_{t,j})} & i \in \boldsymbol{V}_t^- \\ 
0 & \text{otherwise}
\end{cases}
\end{equation}
The learnable portfolio vector $[\boldsymbol{w}^+_t, \boldsymbol{w}^-_t] \in \mathbb{R}^{2N}$ at the $t$-th timestamp satisfies:  
\begin{equation}
\sum_{i=1}^N w^+_{t,i} = 1, \quad \sum_{i=1}^N w^-_{t,i} = -1
\end{equation}  
enforcing capital allocation constraints while maintaining differentiable optimization properties.


\subsection{RL Optimization}
Given the reinforcement learning formulation of portfolio management, we optimize the investment policy $\pi_\theta(a|s)$ through policy gradient methods, where the action $a = [\boldsymbol{w}^+, \boldsymbol{w}^-]$ represents portfolio weights generated from market state $s = \{X^{macro}, X^{micro}\}$. The policy $\pi_\theta$ defines a probability distribution over trading trajectories $\tau = \{s_1,a_1,...,s_T,a_T\}$ parameterized by $\theta$. Aligned with Equation~\ref{eq_sharpe}, the Sharpe ratio $y_t$ serves as the immediate reward signal quantifying risk-adjusted performance at period $t$. The policy gradient objective maximizes the expected logarithmic Sharpe ratio over trajectories:  
\begin{equation}
J(\theta) = \mathbb{E}_{\tau \sim \pi_\theta}\left[\sum_{t=0}^{|\tau|} \log(y_t \pi_\theta(a_t|s_t))\right]
\end{equation}

Traditional policy gradient methods suffer from two key limitations: (1) reliance on on-policy data collection requiring expensive environment interactions, and (2) high variance in gradient estimates due to importance weight ratios. Modern approaches like TRPO~\cite{schulman2015trust} and PPO~\cite{schulman2017proximal} address these through constrained optimization and clipped objectives. Specifically, PPO introduces a pessimistic clipping mechanism on the policy ratio:  
\begin{equation}
J_{PPO}(\theta')=\sum_{\tau \sim \pi_{\theta}}\sum^{|\tau|}_{t=0}\text{log}(\text{min}(\frac{\pi_{\theta',t}}{\pi_{\theta,t}}y_t,\text{clip}(\frac{\pi_{\theta',t}}{\pi_{\theta,t}},1\pm \epsilon)y_t) \label{eq_ppo}
\end{equation}
where $\epsilon$ controls the clipping range. This eliminates explicit KL constraints while enabling multiple updates per data batch.  

Our approach modifies the gradient clipping mechanism by applying constraints directly to the policy gradient objective~\cite{markowitz2023clipped}:  
\begin{equation}
J_{Ours}(\theta')=\sum_{\tau \sim \pi_{\theta}}\sum^{|\tau|}_{t=0}\text{min}(\text{log}\pi_{\theta',t}y_t, \text{log}(\text{clip}(\frac{\pi_{\theta',t}}{\pi_{\theta,t}},1\pm \epsilon)\pi_{\theta,t})y_t) \label{COPG}
\end{equation}
This formulation retains two critical advantages: variance reduction and exploration preservation. Our objective function eliminates unbounded importance weights through ratio clipping and maintains policy update diversity via gradient direction constraints. Both mechanisms synergistically mitigate premature convergence to suboptimal policies while ensuring stable learning dynamics in non-stationary financial environments.

\subsection{Trajectory Selection with Ranking Buffer}
To enhance sampling efficiency and mitigate gradient variance, we integrate policy updates with a self-supervised ranking mechanism. The core innovation lies in preferentially retaining high-reward trajectories from policy-environment interactions, thereby creating implicit behavioral priors for policy improvement. The self-supervised module operates through three key principles: trajectory quality assessment using episode-wise Sharpe ratios, rank-aware experience replay with elite trajectory retention, and curriculum learning through iterative policy refinement. This framework treats high-reward trajectories as expert demonstrations, enabling stable policy imitation while preserving exploration capacity. Starting from random policy, we collect some good trajectories and improve the policy with supervised learning; in the next iteration, we sample data with slightly better policy and thus are more likely to find better trajectories. This design allows us to keep the loss function unchanged and slightly modify the batch processing to learn policy. 

\begin{figure}[!tp]
\vspace{-1em}
\begin{algorithm}[H]
 	\caption{Ranking Buffer-enhanced Policy Optimization}
 	\label{alg1}
 	\begin{algorithmic}[1]
 		\State\textbf{Input:} Training steps $\mathcal{S}$, buffer capacity $D$
		\State Initialize policy $\pi_{\theta}$, buffer $\mathcal{D}\gets\emptyset$
 		\For{iteration $=1,2,\ldots$}
 			\State Generate trajectory $\tau\sim\pi_{\theta}$ with Sharpe ratio $y_\tau$
 			\State Store $\{(s_t,a_t,y_\tau)\}_{t=1}^{|\tau|}$ in $\mathcal{D}$
 			\State Rank $\mathcal{D}$ by $y_\tau$ descendingly  
 			\If{$|\mathcal{D}|>D$}
 				\State Truncate $\mathcal{D}$ to top-$D$ trajectories
 			\EndIf
 			\For{step $=1$ to $\mathcal{S}$}
 				\State Sample batch $\mathcal{B}\subset\mathcal{D}$
 				\State Update $\pi_{\theta}$ via $\nabla_\theta J_{Ours}$ (Eq.~\ref{COPG})
 			\EndFor
 		\EndFor
  	\end{algorithmic}
 \end{algorithm}
\vspace{-2em}
\end{figure}

As detailed in Algorithm~\ref{alg1}, the financial market's stochasticity enables sporadic discovery of high-reward trajectories even with suboptimal policies. The ranking buffer $\mathcal{D}$ maintains an evolving elite set of trajectories through: discarding trajectories below percentile thresholds; preventing mode collapse through stochastic sampling; gradually raising performance benchmarks.

This design guarantees policy updates derive exclusively from substantively informative trajectories, effectively filtering out noisy/low-yield samples that hinder convergence. The constant inflow of elite demonstrations creates a virtuous cycle - improved policies generate better trajectories, which in turn refine the policy through more targeted imitation.

 \begin{table}[!bp]
	\centering
	\vspace{-15pt}
	\caption{Statistics of the experimental datasets.}
	\label{tab_data}
	\resizebox{0.9\columnwidth}{!}{
	\begin{tabular}{c|c|c|c}
	\toprule
	Index & \#Stocks & Training & Test\\
	\midrule
	CSI 300 & 300 & 2014.01-2021.12 & 2022.01-2024.04\\
	CSI 500 & 500 & 2014.01-2021.12 & 2022.01-2024.04\\
	CSI 1000 & 1000 & 2014.01-2021.12 & 2022.01-2024.04\\
	\bottomrule
	\end{tabular}}
\end{table}

\section{Experiments}
We conduct rigorous empirical validation through three methodological dimensions. Firstly, evaluating SCP's risk-adjusted returns (Sharpe ratio, maximum drawdown) against state-of-the-art baselines across three financial indices. Subsequently, quantifying individual contributions of core technical innovations through controlled removal experiments. Finally, systematic investigation of model robustness to critical configuration parameters.

\subsection{Experimental Settings}
\textbf{Dataset Description.} Our index enhancement strategy employs constituent stocks from three benchmark indices in China's A-share market: CSI 300 (large-cap), CSI 500 (mid-cap), and CSI 1000 (small-cap). Table~\ref{tab_data} summarizes the key statistical characteristics of these indices. While Chinese retail investors face regulatory restrictions on direct short selling, we evaluate both long-only and long-short portfolio configurations in simulated environments to comprehensively assess model capability. For real-world deployment, the framework operates exclusively through long position allocations determined by the model's weight predictions.

\begin{table*}[!tp]
	\centering
	\label{all_ls_res}
	\caption{Overall performance comparison with the long-short strategy.}
	\label{tab_all_res}
	\resizebox{1\linewidth}{!}{
	\begin{tabular}{c|cccccc|cccccc|cccccc}
	\toprule
	\multirow{2}{*}{Model} & \multicolumn{6}{c|}{CSI300} & \multicolumn{6}{c|}{CSI500} & \multicolumn{6}{c}{CSI1000}\\
	\cmidrule(l{8pt}r{8pt}){2-7}\cmidrule(l{8pt}r{8pt}){8-13}\cmidrule(l{8pt}r{8pt}){14-19}
	& ARR(\%)$\uparrow$ & AVol$\downarrow$ & ASR$\uparrow$ & MDD(\%)$\downarrow$ & CR$\uparrow$ & SoR$\uparrow$ & ARR(\%)$\uparrow$ & AVol$\downarrow$ & ASR$\uparrow$ & MDD(\%)$\downarrow$ & CR$\uparrow$ & SoR$\uparrow$ & ARR(\%)$\uparrow$ & AVol$\downarrow$ & ASR$\uparrow$ & MDD(\%)$\downarrow$ & CR$\uparrow$ & SoR$\uparrow$\\
	\midrule
	market & -12.03 & 0.17 & -0.71 & 35.34 & -0.34 & -1.19 & -11.59 & 0.20 & -0.58 & 39.35 & -0.29 & -0.92 & -13.95 & 0.23 & -0.60 & 46.22 & -0.30 & -0.91\\
	BLSW & -2.08 & 0.07 & -0.30 & 11.84 & -0.18 & -0.50 & 0.92 & 0.06 & 0.15 & 10.17 & 0.09 & 0.26 & 3.66 & 0.06 & 0.57 & 8.13 & 0.45 & 1.00\\
	CSW & 1.67 & 0.07 & 0.24 & 8.89 & 0.19 & 0.39 & -1.29 & 0.06 & -0.20 & 10.80 & -0.12 & -0.33 & -3.76 & 0.06 & -0.58 & 14.37 & -0.26 & -0.93\\
	\midrule
	LightGBM & -1.53 & \textbf{0.04} & -0.39 & 6.99 & -0.22 & -0.69 & 7.16 & \textbf{0.04} & 1.49 & \textbf{4.69} & \textbf{1.52} & \textbf{2.91} & 10.57 & 0.09 & 1.21 & 6.13 & 1.73 & 2.34\\
	DA-RNN & 0.34 & 0.06 & 0.06 & 12.02 & 0.03 & 0.11 & 7.16 & 0.07 & 1.02 & 5.47 & 1.31 & 2.71 & 11.11 & \textbf{0.05} & \textbf{2.47} & \textbf{3.53} & 3.15 & 4.78\\
	\midrule
	EIIE & -1.35 & 0.07 & -0.19 & 9.60 & -0.14 & -0.34 & -6.46 & 0.08 & -0.77 & 15.94 & -0.41 & -1.14 & 0.78 & 0.08 & 0.10 & 10.38 & 0.08 & 0.16\\
	AlphaStock & -0.43 & 0.05 & -0.09 & 5.96 & -0.07 & -0.14 & -1.17 & 0.05 & -0.24 & 6.09 & -0.19 & -0.43 & 1.96 & 0.06 & 0.34 & 5.60 & 0.35 & 0.54\\
	DeepTrader & 0.46 & 0.08 & 0.05 & 9.61 & 0.05 & 0.09 & 3.64 & 0.04 & 0.88 & 4.87 & 0.75 & 1.62 & 2.06 & 0.07 & 0.29 & 10.46 & 0.20 & 0.52\\
	MetaTrader & 0.79 & 0.13 & 0.06 & 14.47 & 0.06 & 0.11 & 5.56 & 0.05 & 1.03 & 4.74 & 1.17 & 1.99 & 8.93 & 0.13 & 0.71 & 12.88 & 0.69 & 1.45\\
	\midrule
	SCP & \textbf{18.07} & 0.09 & \textbf{1.97} & \textbf{4.76} & \textbf{3.80} & \textbf{4.63} & \textbf{7.99} & 0.05 & \textbf{1.60} & 5.93 & 1.35 & 2.19 & 
	\textbf{19.61} & 0.08 & 2.43 & 4.96 & \textbf{3.95} & \textbf{5.46}\\
	\bottomrule
	\end{tabular}}
\end{table*}

\textbf{Baselines.} 
Our experimental evaluation compares against three categories of baseline methods. Traditional Portfolio Strategies includes Market(Buy-and-hold benchmark portfolio), BLSW~\cite{poterba1988mean}(Mean-reversion strategy buying recent underperformers and selling outperformers), and CSM~\cite{efficiency1993returns}(Cross-sectional momentum strategy exploiting inter-asset return persistence). Deep Supervised Learning Models contains LightGBM~\cite{ke2017lightgbm}(Gradient-boosted decision tree ensemble with automatic feature engineering), DA-RNN~\cite{qin2017dual}(Dual-attention LSTM network capturing temporal and feature-wise dependencies). Deep Reinforcement Learning Frameworks consist of EIIE~\cite{jiang2017deep}(Portfolio optimization via adversarial imitation learning), AlphaStock~\cite{wang2019alphastock}(Industry-aware asset allocation with graph neural networks), DeepTrader~\cite{wang2021deeptrader}(Macro-micro integrated trading policy with risk-aware rewards), and MetaTrader~\cite{niu2022metatrader}(Meta-learning framework for fast market regime adaptation).

\textbf{Evaluation Metrics.} We employ six established portfolio evaluation metrics across three dimensions: 1) return characteristics including Annualized Return Rate (ARR); 2) risk profile including Annualized Volatility (AVol), Maximum Drawdown (MDD); 3) risk-ddjusted Returns including Annualized Sharpe Ratio (ASR), Calmar Ratio (CR), Sortino Ratio (SoR). Metric optimization directions follow standard financial conventions: minimize AVol and MDD while maximizing ARR, ASR, CR, and SoR. 

\textbf{Parameter Setup.} We implemented our model using PyTorch and meticulously tuned the key parameters. The feature hidden dimension is $d=32$. The lengths of the look-back window for the daily and intraday branches are 20 days and 5 days respectively. The max steps of reinforcement learning are set to 20. Top-$M$ is set to 50\% and 20\% of total numbers of index component assets in the long-short and only long strategy respectively. In the follow-up ablation subsection, we carefully fine-tuned the layers $L_1, L_2$ of temporal and spatial transformers, the buffer size $D$ of ranking trajectories, and the learning rate $lr$ to select the optimal hyper-parameters. Regarding the more detailed settings of the experiment, we will release the source code of the SCP later.

\subsection{Performance Comparison}
Table \ref{all_ls_res} presents a comparative analysis of our SCP model against three baseline categories under long-short trading conditions, yielding three principal findings:
\begin{itemize}
	\item The proposed model demonstrates statistically significant outperformance across all three benchmark indices (CSI 300/500/1000), achieving optimal balance between risk-adjusted returns and portfolio stability. Compared to the state-of-the-art MetaTrader framework, SCP exhibits 32×, 0.5×, and 2× improvements in Annualized Sharpe Ratio (ASR) respectively, validating the effectiveness of integrating micro-level market microstructure signals with self-supervised clipped-objective optimization. This performance differential highlights the criticality of reinforcement learning architectures in processing high-dimensional financial time series while maintaining risk constraints.
	\item Traditional strategies exhibit conditional efficacy - while the CSM method outperforms Market and BLSW baselines on CSI 300, its performance degrades to sub-baseline levels on CSI 500 and CSI 1000. Deep supervised baselines such as LightGBM and DA-RNN struggle in risk criterion, as they only consider accurate prediction of price movements while ignoring the costly consequences of inaccurate predictions. Deep reinforcement learning approaches consistently outperform traditional and deep supervised methods in achieving a balance between risk and return. However, RL-based methods are sensitive and fragile, making it challenging to fine-tune parameters for optimal performance.
	\item Notably, CSI 1000 implementations show constrained optimization potential compared to CSI 300/500 counterparts. We hypothesize this stems from two structural factors. First, the CSI 1000's broader constituent base (1,000 mid-cap stocks vs 300/500 large-caps) increases state-space complexity for RL-based portfolio optimization. And higher constituent turnover rates in mid-cap indices create temporal discontinuity in policy learning.
\end{itemize}

\begin{figure*}[!tp]
	\centering
	\includegraphics[width=1\linewidth]{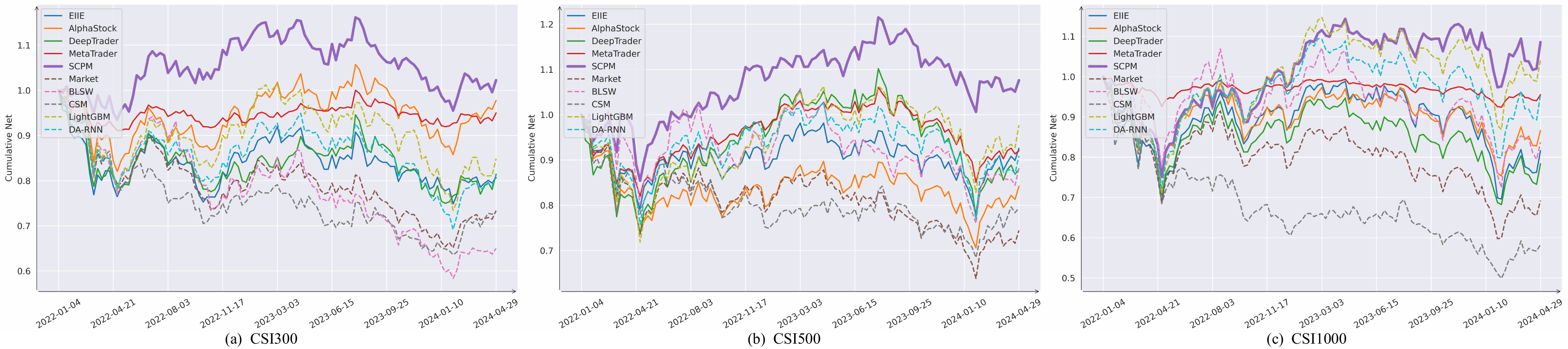}
	\vspace{-20pt}
	\caption{The cumulative net worth performance of ours and benchmarks across datasets with the long top-20\% position strategy.}
	\vspace{-15pt}
	\label{all_l_res}
\end{figure*}

Figure~\ref{all_l_res} demonstrates the cumulative capital trajectories under long-only configurations, yielding two principal findings:
\begin{itemize}
	\item SCP consistently demonstrates superior performance across three index datasets. Specifically, it achieves significant improvements over the strongest baseline method (e.g., MetaTrader) on both SCI300 and SCI500 in terms of both profitability and risk control. This is enabled by our self-supervised clipped-objective policy learning framework, which ensures stability in portfolio generation. Consequently, our model addresses the common instability challenges inherent in RL-based optimization.
	\item In long-only investment scenarios, traditional strategies exhibit suboptimal accumulated returns, primarily due to their insufficient perception of market dynamics and inability to forecast future asset trends. This limitation stems from two fundamental issues: (1) Supervised learning methods, while establishing asset return-weight allocation mappings through nonlinear regression, create indirect correlations that overlook portfolio risk equilibrium; (2) Although RL-based approaches demonstrate advantages in Sharpe ratio and maximum drawdown metrics compared to supervised baselines, their cumulative returns remain constrained. Our framework bridges this gap by capturing high-frequency intraday market microstructure through adaptive temporal modeling, combined with novel risk-sensitive reward mechanisms during policy optimization. This synergistic architecture achieves state-of-the-art performance across both return generation and risk mitigation dimensions.
\end{itemize}

\subsection{Ablation Studies}
\begin{table*}[!tp]
	\centering
	\caption{Ablation studies of self-supervised clipped-objective portfolio management with the long-short strategy.}
	\label{tab_ablation}
	\resizebox{1\linewidth}{!}{
	\begin{tabular}{c|cccccc|cccccc|cccccc}
	\toprule
	\multirow{2}{*}{Model} & \multicolumn{6}{c|}{CSI300} & \multicolumn{6}{c|}{CSI500} & \multicolumn{6}{c}{CSI1000}\\
	\cmidrule(l{8pt}r{8pt}){2-7}\cmidrule(l{8pt}r{8pt}){8-13}\cmidrule(l{8pt}r{8pt}){14-19}
	& ARR(\%) & AVol & ASR & SoR & MDD(\%) & CR & ARR(\%) & AVol & ASR & SoR & MDD(\%) & CR & ARR(\%) & AVol & ASR & SoR & MDD(\%) & CR \\
	\midrule
	Daily PG & 8.16 & 0.10 & 0.85 & 10.17 & 0.80 & 1.63 & 1.86 & 0.08 & 0.22 & 9.04 & 0.21 & 0.43 & 5.65 & 0.10 & 0.54 & 9.32 & 0.61 & 1.08\\
	Two-level PG & 11.80 & 0.11 & 1.04 & 9.30 & 1.27 & 1.90 & 4.53 & 0.08 & 0.59 & 7.24 & 0.63 & 1.17 & 7.93 & 0.09 & 0.84 & 8.79 & 0.90 & 1.56\\
	\midrule
	PPO & 12.81 & 0.08 & 1.54 & 5.79 & 2.21 & 3.25 & 5.00 & 0.07 & 0.69 & 6.75 & 0.74 & 1.23 & 10.02 & 0.08 & 1.24 & 5.66 & 1.77 & 2.54\\
	Clipped PG & 14.23 & 0.09 & 1.57 & 5.91 & 2.41 & 3.41 & 7.37 & 0.08 & 0.96 & 4.45 & 1.65 & 1.74 & 13.98 & 0.05 & 2.90 & 4.06 & 3.45 & 5.55\\
	SCP & 18.07 & 0.09 & 1.97 & 4.76 & 3.80 & 4.63 & 7.99 & 0.05 & 1.60 & 5.93 & 1.35 & 2.19 & 19.61 & 0.08 & 2.43 & 4.96 & 3.95 & 5.46\\
	\bottomrule
	\end{tabular}}
	\vspace{-5pt}
\end{table*}

To systematically assess the contributions of SCP's components, we conduct ablation studies across three datasets. The complete SCP framework is compared against four variants: (1) Daily-PG (baseline with macro-feature encoder + policy gradient), (2) Two-Level-PG (adds micro-market status modeling), (3) PPO-SCP (replaces policy update mechanism), and (4) Clipped-Objective-PG (modifies loss function), as quantitatively analyzed in Table \ref{tab_ablation}. Three key findings emerge:
\begin{itemize}
	\item The model's effectiveness stems from four synergistic innovations: intraday asset representation via temporal convolution networks, proximal policy optimization (PPO) framework, clipped-objective reward shaping, and self-supervised trajectory quality-aware filtering.
	\item As demonstrated in line 1 and line 2, integrating micro-level market information that captures intraday asset volatility enhances metrics like the Sharpe ratio. This approach introduces a new framework for translating high-frequency data into daily asset trend predictions.
	\item As demonstrated in line 3, line 4, and line 5, the adaptation of PPO learning with a clipped-objective reward function proves crucial in mitigating the risk of getting stuck in local optima. Additionally, the self-supervised trajectory ranking buffer enhances training by providing high-quality samples, thereby accelerating model convergence. This component acts as a module for policy gradient reinforcement learning algorithms, facilitating the generation of more precise and stable strategies.
\end{itemize}

\subsection{Parameter Analysis}
In this subsection, we systematically investigate three core hyperparameters through controlled experiments. First, the role of stacked Transformer layers in joint temporal encoders and spatial graph attention networks (GATs). Second, the regulatory effect of ranking buffer size on low-reward trajectory filtration efficiency. Third, the sensitivity of learning rate to off-policy gradient stability under non-stationary market perturbations.
\begin{figure*}[!tp]
	\centering
	\includegraphics[width=1\linewidth]{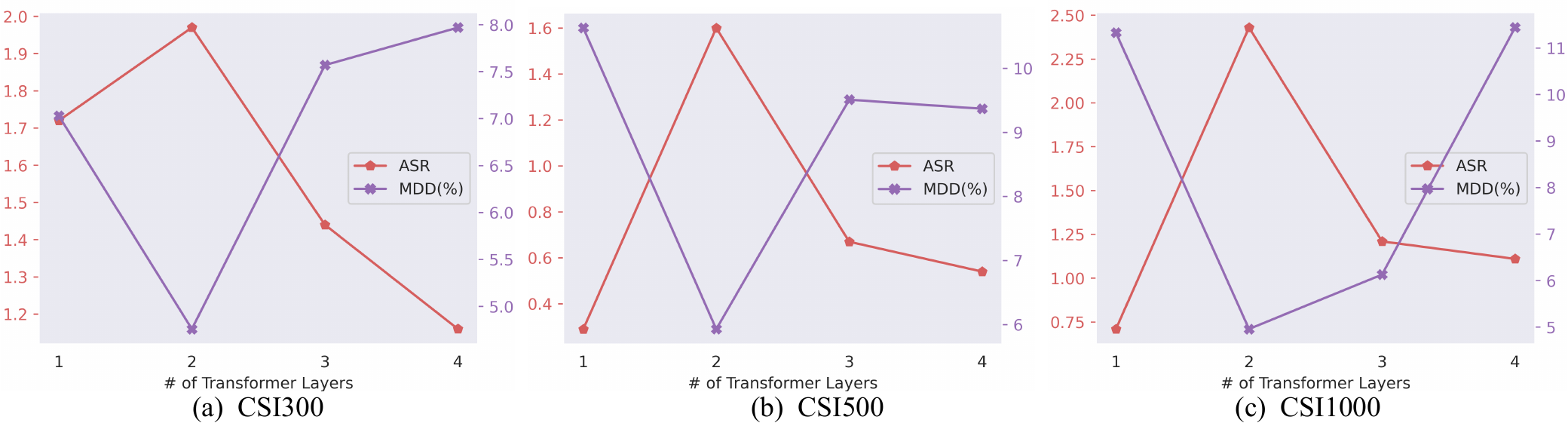}
	\vspace{-20pt}
	\caption{Performances comparison of setting different Transformer layers $L$.}
	\vspace{-5pt}
	\label{fig_layer}
\end{figure*}

\textbf{Impact of $L$.}  We evaluate SCP's performance with temporal layers $L$ ranging from 1 to 4 across three datasets. Due to GPU resource constraints and training efficiency requirements, we maintain relatively shallow Transformer layers. Figure \ref{fig_layer} shows that performance first improves then declines as $L$ increases. The peak performance at $L=2$ suggests that: With $L<2$, temporal relationships are not fully exploited; When $L>2$, the model learns less consistent temporal patterns, which negatively impacts performance. Therefore, we set $L=2$ in the Transformer architecture to optimally capture asset time-series trends.


\begin{figure}[!tp]
	\centering
	\includegraphics[width=0.8\columnwidth]{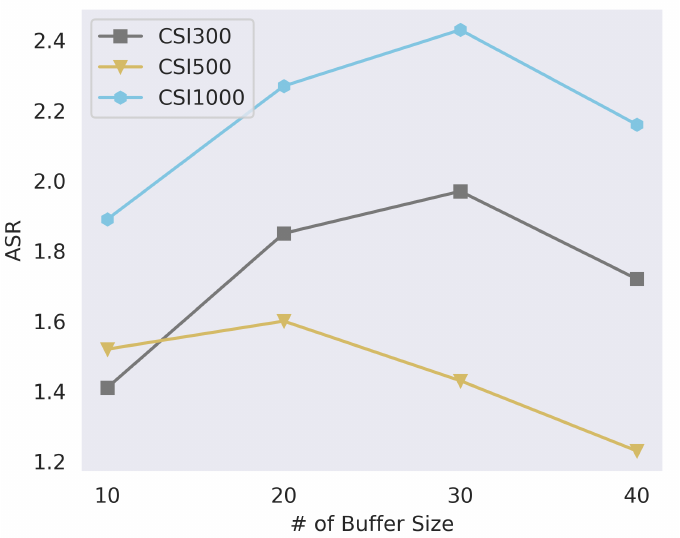}
	\vspace{-10pt}
	\caption{Performances with various buffer sizes $D$.}
	\vspace{-10pt}
	\label{fig_buffer}
\end{figure}

\textbf{Impact of $D$.} 
The buffer size $D$ plays a critical role in our proposed trajectory filtration mechanism, where we maintain a fixed-size buffer to enhance policy learning efficiency. For implementation consistency, we set the training batch size to half of $D$, guaranteeing that each parameter update replaces 50\% of stored trajectories. We investigate the impact of different buffer sizes $D$ within the range $\{10,20,30,40\}$. From the results illustrated in Figure \ref{fig_buffer}, we observe significant variations in optimal buffer sizes across datasets. Specifically, $D = 30, 20, 30$ are found to be the best settings for CSI 300, CSI 500, and CSI 1000, respectively. Increasing $D$ allows the model to compare more high-quality samples in each mini-batch, facilitating the exploration of similar strategies. However, enlarging $D$ can lead to inconsistencies in sampled trajectories over time due to dataset scale and portfolio asset numbers, potentially resulting in degraded model performance. This observation emphasizes the importance of carefully selecting $D$ to optimize our framework's performance effectively.

\begin{figure}[!tp]
	\centering
	\includegraphics[width=0.8\columnwidth]{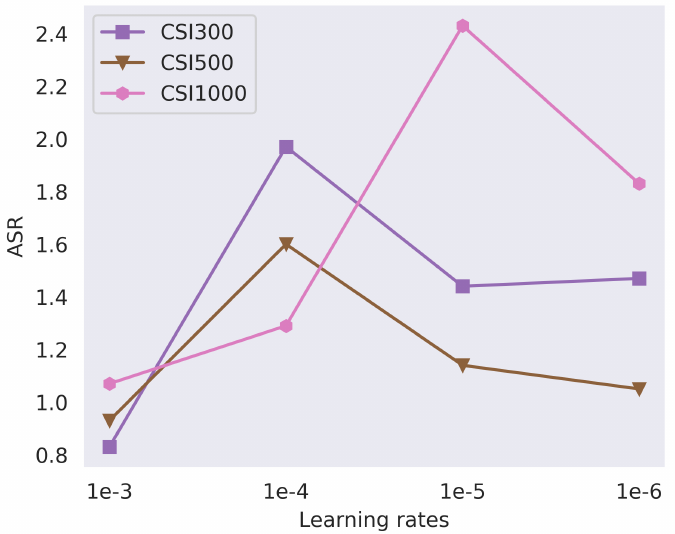}
	\vspace{-10pt}
	\caption{Statistics of learning rates on three datasets.}
	\vspace{-10pt}
	\label{fig_lr}
	\
\end{figure}

\textbf{Impact of $lr$.} 
In selecting an optimizer for our model, we compared naive SGD, RMStep, Adam, and found that Lion~\cite{chen2024symbolic} performs exceptionally well. Lion optimizer is noted for its memory efficiency and strong generalization capabilities across different architectures, datasets, and tasks. It also exhibits relative insensitivity to learning rate adjustments in deep reinforcement learning settings. We conducted experiments with different learning rates $lr$ from $\{1e^{-3},1e^{-4},1e^{-5},1e^{-6}\}$ with $L2$ regularization, evaluated across three datasets, and present the results in Figure \ref{fig_lr}. Our findings indicate that a smaller learning rate ensures stable training of the model and effectively supports the application of the clipped-objective loss function. Specifically, $lr= 1e^{-4}, 1e^{-4}, 1e^{-5}$ emerge as suitable choices for CSI300, CSI500, and CSI1000, respectively. In conclusion, our investigations underscore the importance of this hyperparameter in SCP, highlighting its role in fine-tuning the gradient descent weights for policy learning with flexibility and efficiency.

\section{Conclusions and Future Work}
This study presents a self-supervised clipped-objective portfolio management framework (SCP) that synergistically integrates daily and intraday market signals to construct enhanced asset status representations, thereby capturing the composite drivers of price fluctuations. The framework implements a trajectory ranking buffer that systematically filters suboptimal low-risk-low-return investment paths, optimizing both sample efficiency and computational resource allocation. A clipped-objective off-policy update mechanism further stabilizes the learning process through safeguarded iterative updates that mitigate data variance. Empirical validation across three major financial market datasets demonstrates SCP's superior effectiveness, with comparative experiments and ablation studies systematically quantifying its performance advantages. Future research will focus on integrating large language models (LLMs) with reinforcement learning architectures to enhance computational capabilities for complex financial problems, particularly through multimodal market microstructure analysis and adaptive reward mechanisms for achieving more precise forecasting of price momentum patterns.

\section*{Acknowledgments}
This work has been supported in part by postdoctoral stations of Southwest Securities.

\bibliography{MuZ_scp.bib}

\end{document}